\documentclass[letterpaper, 10 pt, conference]{ieeeconf} 
\IEEEoverridecommandlockouts

\usepackage{times}

\usepackage{cite}
\usepackage{multicol}
\usepackage[bookmarks=true]{hyperref}

\usepackage{graphicx}  
\usepackage{times}  
\usepackage{rotating}  
\usepackage[normalem]{ulem}  
\usepackage{bbold}
\let\labelindent\relax
\usepackage{enumitem}
\usepackage{amssymb}

\usepackage{etoolbox}
\usepackage{caption}
\usepackage{subcaption}
\usepackage{amsmath}
\usepackage{comment}
\usepackage{nicefrac}
\usepackage{amssymb}
\usepackage[makeroom]{cancel}
\usepackage{url}

\usepackage{xcolor}
\definecolor{c1}{RGB}{204,0,204}
\definecolor{c2}{RGB}{204,0,102}
\definecolor{c3}{RGB}{0,0,0}
\definecolor{c4}{RGB}{0,102,204}
\definecolor{c5}{RGB}{0,0,204}
\definecolor{c6}{RGB}{102,0,204}
\usepackage{enumitem}
\usepackage{bbm}
\usepackage{mathtools, cuted}
\usepackage{tabularx}
\newcommand\scalemath[2]{\scalebox{#1}{\mbox{\ensuremath{\displaystyle #2}}}}
\newcommand{\etal}{\textit{et al}. }
\newcommand{\ie}{\textit{i}.\textit{e}., }
\newcommand{\eg}{\textit{e}.\textit{g}. }
\newcommand{\SMone}{0.9}

\newcommand{\Munzir}[1]{{\color{black}{#1}}}
\usepackage{booktabs}
\usepackage{ulem}

\begin{document}

\title{\LARGE \bf
Hierarchical Optimization for Whole-Body \\
Control of Wheeled Inverted Pendulum Humanoids
}

\author{Munzir Zafar, Seth Hutchinson, Evangelos A. Theodorou
\thanks{Munzir Zafar, Seth Hutchinson and Evangelos A. Theodorou are with the Institute of Robotics and Intelligent Machines at the Georgia Institute of Technology, Atlanta, GA, 30332, USA. email: {\tt\small mzafar7@gatech.edu, seth@gatech.edu, evangelos.theodorou@gatech.edu}}
}


%

\maketitle

\begin{abstract}
In this paper, we present a whole-body control framework for \textit{Wheeled Inverted Pendulum (WIP) Humanoids}. WIP Humanoids are redundant manipulators dynamically balancing themselves on wheels. Characterized by several degrees of freedom, they have the ability to perform several tasks simultaneously, such as balancing, maintaining a body pose, controlling the gaze, lifting a load or maintaining end-effector configuration in operation space. The problem of whole-body control is to enable simultaneous performance of these tasks with optimal participation of all degrees of freedom at specified priorities for each objective. The control also has to obey constraint of angle and torque limits on each joint. The proposed approach is hierarchical with a low level controller for body joints manipulation and a high-level controller that defines center of mass (CoM) targets for the low-level controller to control zero dynamics of the system driving the wheels. The low-level controller plans for shorter horizons 
while considering more complete dynamics of the system, while the high-level controller plans for longer horizon based on an approximate model of the robot for computational efficiency. 

\end{abstract}

\IEEEpeerreviewmaketitle

\section{Introduction}

Wheeled inverted pendulum systems offer the best of two worlds. First, their wheels make for inherently fast and efficient locomotion---something that bipedal system designers are still struggling to achieve. Second, the dynamically balancing inverted pendulum endows them with the ability to deal with very heavy payloads, their torques being canceled by a readily adjustable CoM---something that statically stable wheeled platforms can not achieve. These characteristics make them attractive for a wide range of applications such as Segway human transporters \cite{kamen2001segwayPT,solis2009development,tsai2010adaptive,li2012neural,hata2014development}, transporters with seats \cite{hata2014development,huang2010development,petrov2010dynamic,li2012neural,vermeiren2011modeling}, self-balancing wheel chairs \cite{takahashi1999front,takahashi2001back,takahashi2003soft,takahashi2005human} and WIP Humanoids \cite{jeong2007wheeled,sasaki2008pushing,murakami2009motion,acar2011multi,kuindersma2009dexterous,stilman2010golem,
feng2011modeling,fukushima2015sliding,bostondynamics2017handle}. WIP Humanoids add to the abilities of a WIP by offering a redundant manipulator, with one or more arms, that can be controlled to intelligently interact with their environment and perform useful tasks. If they could be controlled for safely handling large forces, their ability to dynamically cancel out their effects would prove useful for assisting humans with tasks requiring large effort.  The WIP humanoid in  \cite{stilman2010golem} was designed with this purpose in mind. Its two-arm-bearing torso is mounted on massive base and spine links, thus providing it with the ability to perform heavy tasks by manipulating its weight torque as needed.

The challenge in controlling such robots is their nonlinear, highly unstable and under-actuated dynamics. A large body of literature exists to deal with these typical problems of WIP systems \cite{grasser2002joe,li2007mechanical,coelho2008development,huang2010development,pathak2005velocity,angeli2001almost,marzi2005multi,qingcheng2014fuzzy,ren2008motion,jung2008control,tsai2010adaptive, wu2011ts,wu2011design,prasad2011optimal,fukushima2015sliding,ahmad2009genetic,goher2009ga,goher2010genetic}. However, the focus of most studies remains a simplified system having one-link attached to the wheels. This has inspired some to leverage this work for WIP humanoid control, by treating the control of wheeled inverted pendulum independently from the control of upper body---the former being a simplified model of the full robot with one link of an equivalent center of mass (CoM). This technique was also utilized  to control the WIP humanoid system in  \cite{stilman2014robots}. 

The problems associated with such a treatment become apparent when precision-critical or safety-critical tasks are performed. For example, since forward motion is left up to the wheels alone, and the upper body is not compensating for pitch changes induced by the demands of locomotion on the wheel 
controller, the end-effector can hardly obey constraints on its position and orientation during locomotion, which may sometimes be critical to perform a task. Similarly, when dealing with sudden changes in large forces, the transients induced on a balancing controller due to the switching of equilibrium positions can not be properly managed if the body is blind to the control of its center of mass. This makes the case for a unified approach to locomotion and manipulation tasks. 

In this work we  propose a unified framework that deals with whole-body control of WIP humanoid structures such as Golem Krang \cite{stilman2010golem}, and we aim to demonstrate the application of the proposed solution on this platform. 
In particular our core contributions are:

{ 
\begin{itemize}
  \item Deriving the full 3D dynamic model of a typical  WIP  Humanoid 
  
  \item Developing a hierarchical control framework in which different tasks can be performed on the WIP platform in a unified fashion resulting in a better overall performance

  \item \Munzir{Showing} how quadratric programming (QP) based control techniques \cite{feng2016online} can be applied as a low-level controller to isolated manipulator dynamics obtained by elimination of wheel dynamics from the dynamic model
  
  \item \Munzir{Analysing} the zero dynamics of the system \Munzir{to motivate the use of a simplified Wheeled Inverted Pendulum Model (WIPM) that approximates the inertial properties of the full robot} for predictive control by a high-level controller
  
  \item Using differential dynamic programming (DDP) and model predictive control (MPC) \Munzir{for generating and controlling CoM trajectory of the WIP Humanoid using predictions of the WIPM. This serves as the high-level controller for providing feasible CoM targets to the low-level controller to ensure effective balancing and locomotion along with the performance of other tasks}
\end{itemize}
}
A brief outline of this document is as follows: 
Section \ref{sec:dyn} describes the dynamic model of the system under consideration. Section \ref{sec:des} gives complete description of the hierarchical approach for whole body control of the robot. Results on a 7-DOF planar robot are presented in section \ref{sec:res}. They represent an illustrative scaled down version of a 19-DOF 3D simulation video submitted with this paper. This is followed by conclusion in section \ref{sec:con}.


\section{Modeling} \label{sec:dyn}
\begin{figure}
 \centering
 \includegraphics[width=0.6\columnwidth]{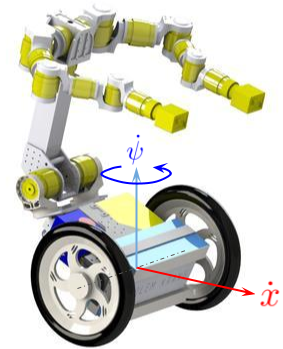}
 \caption{A Wheeld Inverted Pendulum Humanoid. 3D Simulation video provided with this paper applies the proposed approach on this robot.}
 \label{fig:frames}
\end{figure}
In this section, we present the dynamic model of a WIP humanoid under consideration  \cite{stilman2010golem}. The techniques usually utilized for dynamic modeling of a WIP system are Euler-Lagrange, Newton's laws of motion or Kane's method. Assumptions include rigid bodies, flat ground and zero slip. Another simplification for dynamic modeling is to restrict motion to the sagittal plane \cite{akesson2006design, ha1994trajectory, jian2008research, kim2011dynamic, li2007mechanical, lien2006implementation, nomura2009adaptive, ooi2003balancing}. Our analysis does not impose this restriction, and our robot is allowed to explore the full range of accessible workspace. Others have explored this approach \cite{grasser2002joe, hu2008design, takei2009baggage, tsai2007pilot, wu2011robust}; however, they perform their analysis on simple WIP systems, with the body consisting of either just one link, or a simplifying assumption is used to treat it as such. Our analysis does not make any such simplifying assumptions in the derivation of the model.

The system under consideration is shown in Figure \ref{fig:frames}. It is a highly redundant manipulator mounted on a differential wheel drive able to dynamically balance itself in an inverted pendulum configuration. We are interested to control the robot that is assumed to be always moving on a horizontal ground, thus limiting its DOFs in free space (\ie as opposed to the base floating in the air). However, we begin with defining a set of coordinates that specify all its DOFs that are able to model the configuration when the base is floating in the air. We do this in order to utilize off-the-shelf simulation tools (\eg DART \cite{lee2018dart}) that make use of fast algorithms like Featherstone's \cite{featherstone1984robot} for computation of quantities required by our control algorithm, such as the inertia matrix. We use the term ``body'' to refer to the entire structure of the robot excluding the wheels and the term ``base link'' to identify the first link of the body attached to the wheels. Assuming $n$ is the total number of links in the body, the full configuration of the robot can be represented using $7+n$ coordinates. These include 6 DOFs of the floating base, 2 angular positions of the wheels relative to the base link, and $n-1$ motions of the links in the body relative to their respective parent links. Applying Euler-Lagrange to derive the dynamic model results in
\begin{align}
    A_{full}\ddot{q}_{full} + C_{full}\dot{q}_{full} + Q_{full} = \begin{bmatrix} 0_{1\times6} & \tau^\top \end{bmatrix}^\top
\end{align}
where $q_{full} \in \mathbb{R}^{7+n}$ is the set of generalized coordinates; $\tau \in \mathbb{R}^{1+n}$ is the vector of torques comprising $2$ wheel torques ($\tau_L$ and $\tau_R$) along with $n-1$ body joint torques; $A_{full}$, $C_{full}$ and $Q_{full}$ are respectively the inertia matrix, Coriolis matrix and gravity vector.

Since we consider the case when this robot is moving on a horizontal ground under non-holonomic constraints, the degrees of freedom of the system reduce to $3+n$, which include: (1) robot heading $\dot{x}$ (Fig \ref{fig:frames}), (2) robot spin $\dot{\psi}$ (Fig \ref{fig:frames}), (3) pitch of the base link with respect to the vertical $\dot{q}_1$ (Fig. \ref{system}) and (4) motions of the $n-1$ joints mounted on the base link $\begin{bmatrix} \dot{q}_2 & ... & \dot{q}_{n} \end{bmatrix}^\top$ (Fig \ref{system}). So the minimum set of coordinates are $\mathbf{\dot{q}} = \begin{bmatrix} \dot{x} & \dot\psi & \dot{q}^\top \end{bmatrix}^\top$ where we define $q = \begin{bmatrix} q_1 & ... & q_n \end{bmatrix}^\top$. Define $J_{tf}$ as the transform Jacobian relating full set of coordinates and minimum set of coordinates
\begin{align}
    \dot{q}_{full} = J_{tf} \mathbf{\dot{q}}
\end{align}
The exact expression for $J_{tf}$ is dependent on the choice of reference frames and is trivially found as a function of $q_1$. Space limitation precludes a detailed account.
The dynamic model in minimum set of coordinates can be obtained as \footnote{Note that $\dot{x}$ is a quasi-velocity \cite{ginsberg1998advanced} and therefore requires formulations other than Euler-Lagrange for correct derivation. We have originally derived \eqref{dyn} using Kane's formulation and noticed that the terms that are not captured in the presented treatment have negligible dynamic effect (they are function of $\dot{\psi}^2$). We therefore omitted Kane's analysis for brevity.} 
\begin{align}
    A\mathbf{\ddot{q}} + C\mathbf{\dot{q}} + Q = B\tau  \label{dyn}
\end{align}
where $A=J_{tf}^\top A_{full} J_{tf}$, $C = J_{tf}^\top A_{full} \dot{J}_{tf} \mathbf{\dot{q}} + J_{tf}^\top C_{full}$, $Q = J_{tf}^\top Q_{full}$, $B\tau = J_{tf}^\top\left(\begin{bmatrix} 0_{1\times6} & \tau^\top \end{bmatrix}^\top\right)$. Here $B \in \mathbb{R}^{(n+2)\times(n+1)}$ is the actuation matrix. Its dimensions indicate that there is one less actuator in the system than the total degrees of freedom. $B$ is defined as
\begin{align}
    B = \begin{bmatrix} \bar{B} & 0_{3\times(n-1)} \\ 0_{(n-1)\times2} & I_{n-1}\end{bmatrix}, && 
    \bar{B} = \begin{bmatrix} \tfrac{1}{R} & \tfrac{1}{R} \\ -\tfrac{L}{2R} & \tfrac{L}{2R} \\ -1 & -1 \end{bmatrix}
\end{align}
As for actuation torques, the following points are noteworthy:
\begin{enumerate}
 \item Forward motion $\dot{x}$ is actuated by sum of wheel torques
 \item Spin motion $\dot{\psi}$ is actuated by the difference of wheel torques
 \item The base joint rotation $\dot{q_1}$ is actuated by the reaction torque on wheel motors fixed on the base and driving the wheels
 \item The rest of the joints in the tree structure are actuated by their respective joint torques
\end{enumerate}
\paragraph{Under-actuated System}
Wheel torques $\tau_{L}$ and $\tau_{R}$ control three motions: $\dot{x}$, $\dot{\psi}$ and $\dot{q}_1$. Define 
\begin{align}
    \tau_{1} = -\left(\tau_{L} + \tau_{R}\right) \nonumber \\ 
    \tau_0 = \frac{L}{2R}(\tau_{L} - \tau_{R}) \label{tau01}
\end{align}
$\tau_0$ actuates $\dot{\psi}$, while $\tau_1$ actuates both $\dot{x}$ and $\dot{q}_1$. This is one of two types of under-actuation that can feature in WIP systems. In this type, reaction torque of the wheel motors acts on the base link as the motors are mounted on the link. In the second type, the base link is connected to the wheel cart via a passive joint, so the base link experiences no direct actuation. The first type is easier to build but more difficult to control. Most literature on WIP systems however focuses on the second type of systems (also noted by \cite{guo2014design}). We focus on the first type of systems. 


\section{Hierarchical Control} \label{sec:des}
The specific problem of whole body control of WIP humanoids is addressed by Toshiyuki \etal \cite{sasaki2008pushing,murakami2009motion,acar2011multi}. Their work differs from ours in four important aspects: In the system we consider, the base-link is experiencing reaction torque of the wheels while they consider only the case of a passive base joint. They restrict the angular motion of this joint whereas we allow this joint to participate in body manipulation. Their control treatment is restricted to the sagittal plane, however, we present a framework applicable to 3D manipulations. Finally, their control approach is based on inverse kinematics, as opposed to our formulation where we use inverse dynamics that lends itself easily to compliant motions and force interactions.

Figure \ref{overview} gives an overview of the proposed approach. It is hierarchical with a low level controller responsible for controlling the manipulator/body and a high-level controller that defines center of mass (CoM) targets for the low-level controller to control zero dynamics of the system driving the wheels. The low-level controller plans for shorter horizons while considering more complete dynamics of the system, while the high-level controller plans for longer horizon based on an approximate model of the robot for computational efficiency.


\subsection{Low-Level Controller}
With the definitions of $\tau_0$ and $\tau_1$ in \eqref{tau01} and defining $\Gamma = \begin{bmatrix} \tau_1 & ... & \tau_n \end{bmatrix}$, \eqref{dyn} can be written as
\begin{align}
  A \begin{pmatrix} \ddot x \\ \ddot{\psi} \\ \ddot q \end{pmatrix} + C \begin{pmatrix} \dot x \\ \dot\psi \\ \dot q \end{pmatrix} + Q = \begin{pmatrix} \nicefrac{-\tau_{1}}{R} \\ \tau_0 \\ \Gamma  \end{pmatrix} + \Gamma_{fric},\label{dyn1}
\end{align}
Note that $\Gamma_{fric} \in \mathbb{R}^{(n+1)\times1}$ represents the frictional effects (which we had omitted in previous equations for notational simplicity). In \eqref{dyn1}, we can easily observe that both $\dot{x}$ and $\dot{q}_1$ are actuated by the same torque $\tau_1$.
\subsubsection{Isolating Manipulator Dynamics} \label{sec:iso}
In \eqref{dyn1}, wheel and manipulator dynamics are coupled, not only by the wheel torque $\tau_1$, but also the inertia matrix $A$. In \cite{zafar2016whole}, elimination of $\ddot{x}$ is performed in this system to determine a direct relationship between manipulator joint accelerations $\ddot{q}$ and torques $\Gamma$. The inertia matrix is first defined in block form
\[ A = \begin{bmatrix} a_{xx} & a_{x\psi} & a^\mathsf{T}_{xq} \\ 
    a_{x\psi} & a_{\psi\psi} & a_{\psi q}^\top \\ 
    a_{xq} & a_{\psi q} & A_{qq}  \end{bmatrix}, \]
Then inverting this matrix, followed by some algebraic manipulations, results in the following ODEs that isolate the manipulator dynamics from wheel dynamics
\begin{equation}
\mathcal{A}\ddot q + P\left(C \begin{pmatrix} \dot x \\ \dot q \end{pmatrix} + Q - \Gamma_{fric}\right) = \Gamma, \label{iso3} 
\end{equation}
where:
\begin{align}
 \mathcal{A} &= (I - \beta B)A_{qq}^{*}, ~~ P  = (I - \beta B)\left(\begin{matrix} -\frac{a_{xq}}{a_{xx}}   I \end{matrix}\right),  \nonumber  \\
 A_{qq}^{*} &= A_{qq} - \frac{1}{a_{xx}}a_{xq}a_{xq}^\mathsf{T}, ~~B = \begin{bmatrix} \frac{a_{xq}}{Ra_{xx}} & 0_{n\times(n-1)} \end{bmatrix}.  \nonumber 
\end{align}
The terms $ \beta,\alpha   $  are defined as  $  \beta = \frac{1}{1 + \alpha} \nonumber, \alpha  = \frac{a_{xq1}}{Ra_{xx}}  $  where  $a_{xq1} = \mbox{ first element of } a_{xq} $. The matrix $\mathcal{A}$ is asymmetric. Expression for the new matrices $\mathcal{A}$ and $P$ that appear in \eqref{iso3} are derived in terms of block elements of the original inertia matrix $A$ so there is no need to invert this matrix during run time to find the requried matrices. Using the set of ODEs \eqref{iso3} it is now possible to apply full body control techniques to attain manipulator objectives.

\begin{figure}[h!]
 \centering
 \begin{subfigure}[h]{\columnwidth}
  \includegraphics[width=\columnwidth]{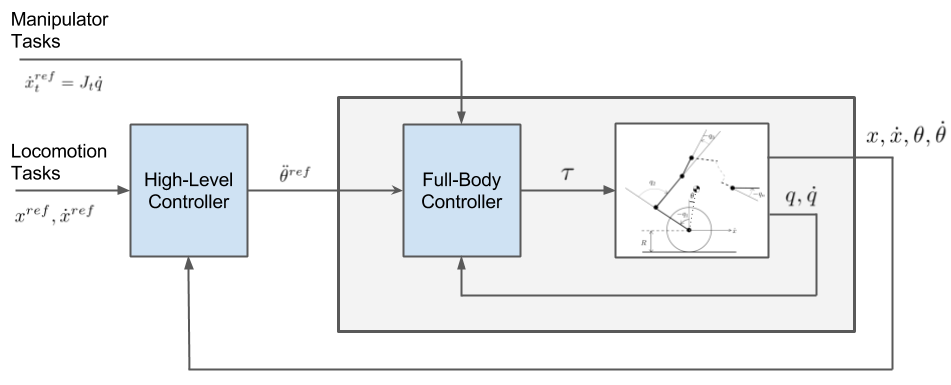}
  \caption{Overview of the proposed approach}
  \label{overview}
 \end{subfigure}  
 \begin{subfigure}[h]{\columnwidth}
  \begin{subfigure}[b]{0.52\columnwidth}
    \centering
    \resizebox{\columnwidth}{!}{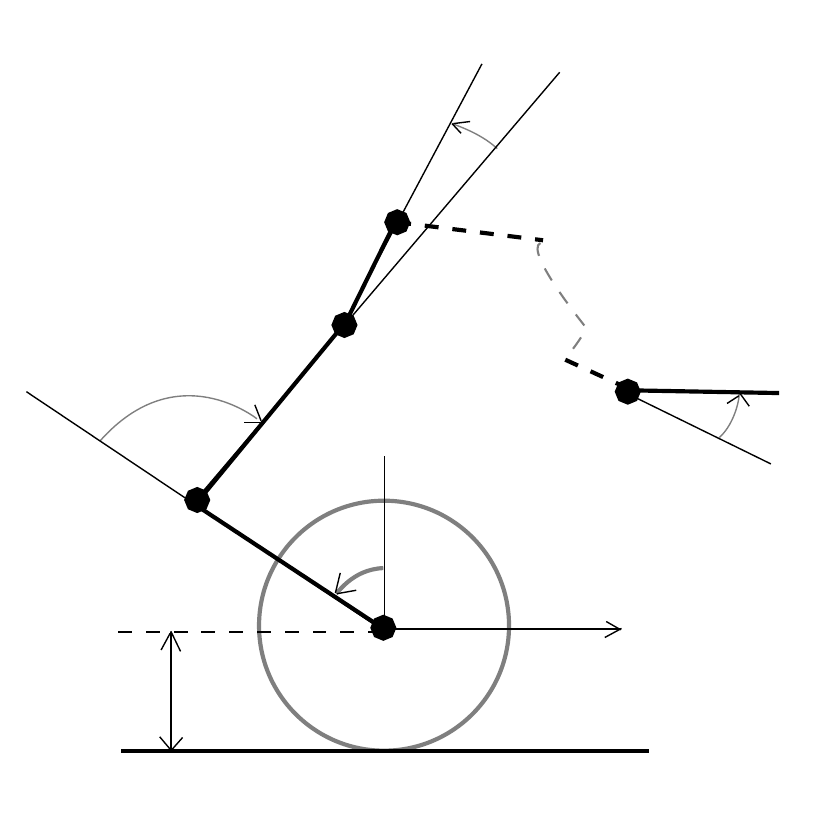}
    \caption{A Wheeled Inverted Pendulum Robot with $n$ joints}
    \label{system}
  \end{subfigure}        
  \begin{subfigure}[b]{0.46\columnwidth}
    \centering
    \resizebox{\columnwidth}{!}{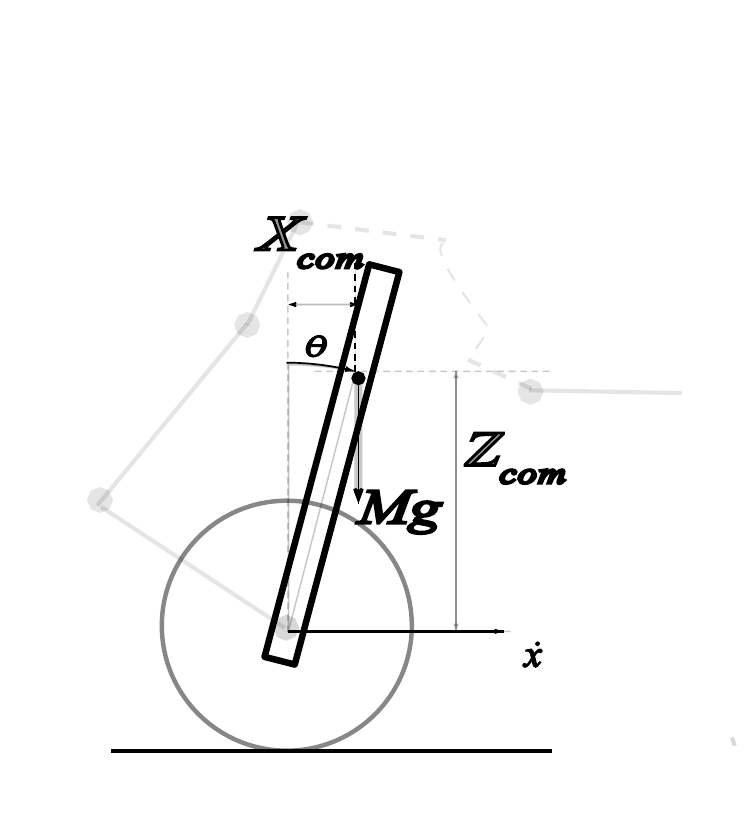}
    \caption{Wheeled Inverted Pendulum Model}
    \label{simple}
  \end{subfigure}
 \end{subfigure}
 \caption{}
\end{figure}


\subsubsection{Optimization Based Control}
The use of inverse dynamics (ID) and inverse kinematics (IK), in conjunction with optimization programs minimizing a single-step cost set up to perform a desired task in the operational space, has gained popularity recently for whole-body control. We prefer the work of \cite{feng2016online} because it offers flexibility in terms of incorporating constraints such as joint, position and torque limits, obstacle avoidance and allows prioritization among tasks. The minimization variables $X$ can be joint torques, accelerations, speeds or contact wrenches. Dynamics are linear in joint accelerations. Similarly, kinematics are linear in joint speeds. Thus constraints involving joint motion will be linear, allowing fast Quadratic Programming (QP) algorithms to solve the optimization. The optimization program is set up as a quadratic cost as follows:
\begin{align*}
 &\min \bigg[ \frac{1}{2} X^\top GX + g^\top X \bigg]\\
& \text{s.t} \: \:  C_{E}X + c_{E} = 0,  ~~ \text{and}~~ C_IX + c_I \leq 0. 
\end{align*}
The cost function takes the form $0.5 \left\|PX -b\right\|^2$ so $G = P^\top P$ and $g = -P^\top b$. An example of cost function designed to perform task $x_t$ (with Jacobian $J_t$), will have $X = \ddot{q}$, $P = J_t$, $b = \ddot{x}_{t}^* - \dot{J_t}\dot{q}$ and $\ddot{x}^*_t = K_p (x_{t}^{ref} - x_t) + K_d(\dot{x}_{t}^{ref} - \dot{x}_t) + \ddot{x}_t^{ref}$. The dynamics of manipulator \eqref{iso3} will be used as the equality constraint to incorporate torque limits. Then perforfming inverse dynamics on \eqref{iso3} with the optimal $\ddot{q}$ obtained from QP will generate the resulting joint torques $\Gamma$. For prioritizing among multiple tasks, $P$ and $b$ can be re-written as \cite{feng2016online}:
\begin{equation}
 P^{\mathsf{T}} = \begin{bmatrix} w_oP_o,  \hdots  w_nP_n \end{bmatrix}, b^{\mathsf{T}}  = \begin{bmatrix} w_ob_o  \hdots  w_nb_n \end{bmatrix}.
\end{equation}

A set of desirable robot behaviors are specified via this cost function as per the goal of the high level controller with higher penalties $w_i$ assigned for higher priority tasks. A summary of the forms $P$ and $b$ matrices will take for various objectives appears in \cite{feng2016online}. 

\subsection{High-Level Controller}
As previously noted, wheel motor torque is also driving the base link. The low-level controller, proposed in the previous subsection, utilizes this torque to directly control only the upper body motions ignoring the wheel dynamics that result from it. This is related to the zero dynamics of the system. The high-level controller proposed in this subsection therefore controls the wheels by planning for body CoM. This is set up as a trajectory optimization problem where a quadratic cost in terms of tracking error is to be minimized. To close the loop for dealing with deviations from optimal path due to unmodeled dynamics, we perform model predictive control (MPC) at sampling intervals $T_s$ for a short horizon $t_H = NT_s$. MPC performs trajectory optimization on a continuous basis over $t_H$ on an updated version of the simplified model, and its first control value is used as the reference for low-level controller QPs over the next sampling period $T_s$.

We use differential dynamic programming (DDP) for trajectory optimization over the full horizon, as well as, optimization in every MPC iteration over the smaller horizon $t_H$. See \cite{jacobson1970differential} for the historical presentation and \cite{tassa2012synthesis} for a modern treatment of DDP. 


\subsubsection{Zero Dynamics}
We now derive the zero dynamics of the full robot and present an analysis motivating the use of CoM trajectory generation for controlling the zero dynamics of the system. We find the zero dynamics of the robot by eliminating $\tau_1$ from the $\dot{x}$ and $\dot{q}_1$ equations in the dynamic model. The two equations of the dynamic model are 
\begin{equation}
 \left(\begin{matrix} a_{xx} & a_{xq}^\top \\ a_{xq} & a_{qq_1}^\top \end{matrix}\right) 
 \left(\begin{matrix} \ddot{x} \\ \ddot{q} \end{matrix}\right) 
 + \left(\begin{matrix} C_x' \\ C_{q_1}'\end{matrix}\right) 
 + \left(\begin{matrix} Q_x \\ Q_{q_1} \end{matrix}\right) 
 = \left(\begin{matrix} -\frac{\tau_1}{R} \\ \tau_1 \end{matrix}\right),
\end{equation}
where $a_{qq_1}$ is the first row of $A_{qq}$, $\left[\begin{matrix} C_x' & C_{q_1}'\end{matrix}\right]^\top$ and  $\left[\begin{matrix} Q_x & Q_{q_1} \end{matrix}\right]^\top$ are the first two elements of $C'$ and $Q$ respectively. Here we have lumped together Coriolis and frictional effects into $C'$ matrix as:
\[
 C' = C\begin{pmatrix} \dot{x} \\ \dot{q} \end{pmatrix} - \Gamma_{fric}.
\]
Eliminating $\tau_1$ from the two equations results in:
\scriptsize
\begin{align}
 &(R a_{xx} + a_{xq_1}) \ddot{x} + \left(Ra_{xq}^\top + a_{qq_1}^\top\right)\ddot{q} + \left( RC_x'+C_{q_1}'\right) + \left(RQ_x+Q_{q_1}\right) = 0 \nonumber 
\end{align}
\begin{align}
 &\Rightarrow 
 \scalemath{1}{\left[ R \left(2 m_w + \frac{2I_w}{R^2} + M\right) + MZ_{com} \right]} \ddot{x} \nonumber  \\
 &+ \scalemath{1}{\left[ R \begin{pmatrix} MZ_{com} \\ M_2 Z_{com(2)} \\ \vdots \\ M_n Z_{com(n)} \end{pmatrix}^\top
 	      + \begin{pmatrix} I   \\ \beta_2    \\ \vdots \\ \beta_n    \end{pmatrix}^\top
	      \right]}\ddot{q} + \left( RC_x'+C_{q_1}'\right) -  MgX_{com} = 0, \label{Fullzero}
\end{align}
\normalsize
where: \begin{itemize}[label={},leftmargin=1.5cm] 
       \item[$m_w$, $I_w$:] are wheel mass and inertia about wheel-axis
       \item[$M$:] is full body mass.
       \item[$I$:] is full body inertia around wheel axis.
       \item[$\scalemath{0.75}{\begin{bmatrix} X_{com},  Z_{com} \end{bmatrix}}:$] are coordinates of body CoM.
       \item[$M_j$:]  is the mass of the articulated structure on joint $i$ defined as $ M_ =\sum_{k=j}^n m_k$. 
       \item[$Z_{com(j)}$:] is the $z$ coordinate of CoM of the articulated structure about the wheel axis.
       \item[$\beta_j$:]  capture coupling torques on wheel axis due to the motion of the articulated structure on joint $i$ 
      \end{itemize}
In deriving \eqref{Fullzero}, we have utilized expressions for $a_{xx}$, $a_{xq_1}$,$a_{xq}$, $a_{qq_1}$, $Q_x$ and $Q_{q_1}$ obtained by symbolic evaluation of \eqref{dyn1} while neglecting effects of spin motion ($\dot\psi$ and $\ddot\psi$) (space limitations preclude a more detailed account). Note that in the absence of transient dynamic forces (\ie when $\ddot{q}=\dot{q} = 0$) the only term driving $\dot{x}$ motion is the torque due to body weight $MgX_{com}$. Modeling the CoM motion ($\dot\theta$, $\ddot\theta$), does capture some effects of the joint motions ($\dot{q}$, $\ddot{q}$). This provides a strong motivation to approximate the full robot as a simplified model---a single link of an equivalent CoM and inertia attached to the wheels---for the purpose of fast trajectory planning to dictate the zero dynamics of the robot for controlling wheel dynamics.


\subsubsection{Simplified Model}
The high-level planner we propose sees the robot as a simplified model that approximates the overall dynamics. This approach is similar to \cite{feng2016online}, where they approximate the bipedal humanoid with a Linear Inverted Pendulum Model (LIPM). We propose to use a Wheeled Inverted Pendulum Model (WIPM). The WIPM is a one-link robot dynamically balancing itself on the wheel (Figure \ref{simple}). Symbolic evaluation of \eqref{dyn} for the case of single link in the body gives us WIPM dynamics. We use $\theta$ to represent the pitch of the link. For the full robot, $\theta$ is the angular position of the body CoM. Using $\dot{x}$ and $\dot{q}_1$ equations in \eqref{dyn}, the model of WIPM is:

\begin{align}
 \scalemath{\SMone}{R\Big(m_w + \frac{I_w}{R^2} + M \Big)\ddot{x} + RMZ_{com}\ddot{\theta} + RMX_{com}\dot{\theta}^2} &\scalemath{\SMone}{= -\tau_{1}} \label{simple1} \\
 \scalemath{\SMone}{MZ_{com}\ddot{x} + I\ddot{\theta} - MX_{com}g} &\scalemath{\SMone}{= \tau_{1}}. \label{simple2} 
\end{align}

The low-level controller described earlier is used to control $\theta$. High-level controller is designed to control the horizontal motion $\dot{x}$ using $\theta$. So it plans for a trajectory of $\theta$ in order to control $\dot{x}$. Eliminating $\tau_1$ from \eqref{simple1} and \eqref{simple2} gives us the zero dynamics of the simplified model:
\begin{align}
 \Big[R\Big(m &+ \frac{I_w}{R^2} + M\Big) + MZ_{com}\Big]\ddot{x} + \left(MZ_{com} + I\right)\ddot{\theta} \nonumber \\ 
 &+ RMX_{com}\dot{\theta}^2 - MgX_{com} = 0. \label{WIPzero}
\end{align}
To obtain the simplified model from the full model, parameters $R$, $m_w$, $I_w$, $x$ and $\dot{x}$ will be directly used. Mass of the full robot is found using $M = \sum_{i=1}^n m_i$ and remains constant. Other parameters $\lambda(q) = \begin{pmatrix} X_{com} & Z_{com} & \theta & \dot\theta & I \end{pmatrix}$ are function of full state $q$ and will be obtained as follows:
\begin{align}
 & \begin{bmatrix} X_{com} \\ Z_{com} \end{bmatrix} = \frac{1}{M} \sum_{i=1}^n m_i \begin{bmatrix} x_{com(i)} \\ z_{com(i)} \end{bmatrix}, ~~ \theta = \arctan\left(\frac{X_{com}}{Z_{com}}\right)\nonumber \\
 &  \dot\theta = \frac{\cos\theta}{Z_{com}} \begin{bmatrix} \cos\theta & -\sin\theta \end{bmatrix} J_{com} \dot{q}, ~~ I = a_{(q1)(q1)}, \nonumber
\end{align}
where $x_{com(i)}$ and $z_{com(i)}$ are CoM coordinates of link $i$, $a_{(q1)(q1)}$ refers to the diagonal element of inertia matrix $A$ corresponding to the generalized acceleration $\ddot{q_1}$, $J_{com}$ is the Jacobian of the CoM.

\subsubsection{Receding Horizon Control}
Dividing \eqref{WIPzero} by $ML$ (where we defined $L = \sqrt{X_{com}^2 + Z_{com}^2}$, $X_{com} = L\sin{\theta}$ and $Z_{com} = L\cos{\theta}$) we get:
\begin{equation}
 \left(\alpha + \cos{\theta} \right)\ddot{x} + \left(\beta + \cos{\theta} \right)\ddot{\theta} = g\sin{\theta} - R\sin{\theta}\dot{\theta}^2
\end{equation}
where $\alpha=\frac{R}{ML}\left(M + m + \frac{I_w}{R^2}\right)$ and $\beta=\frac{I}{ML}$. Next by defining $X = [\begin{matrix} \theta & \dot{\theta} & x & \dot{x} \end{matrix}]^\top$ and $u=\ddot{\theta}$, we can rewrite the dynamics as:
\begin{equation}
\dot{X} = \begin{bmatrix} \dot{\theta} \\ u \\ \dot{x} \\ \frac{g\sin{\theta} - R\sin{\theta}{\dot{\theta}}^2 - \left(\beta + \cos{\theta}\right)u}{\alpha + \cos{\theta}} \end{bmatrix} \label{WIPdyn} = f_c(X,u).
\end{equation}
For allowing the high-level controller to plan spin motion, $\ddot\psi$ equation derived for the single link case from \eqref{dyn} will be added to the above. In that case, the control input is two-dimensional \ie $u=\begin{bmatrix} \ddot\theta_{ref} & \tau_0 \end{bmatrix}^\top$, and we add four more states $\psi$, $\dot\psi$, $X_0$ and $Y_0$ to the state vector, where $(X_0, Y_0)$ is the location in the ground plane of the midpoint between the wheels wrt an inertial frame. $\psi$ and $\dot\psi$ will evolve according to the $\ddot\psi$ equation of the WIPM. While $X_0$ and $Y_0$ will evolve according to $\dot{X}_0 = \dot{x} \cos{\psi}$ and $\dot{Y}_0 = \dot{x} \sin{\psi}$.

Assuming small time-step $\Delta t$ the discretized dynamics of the system can be approximated as
\begin{equation}
 X_{i+1} = X_i + \Delta t f_c(X_i, u_i) = f(X_i, u_i). \nonumber
\end{equation}
Given a desired goal position $X^{ref} = \begin{bmatrix} \theta^{ref} & \dot\theta^{ref} & x^{ref} & \dot{x}^{ref} \end{bmatrix}^\top = \begin{bmatrix} 0 & 0 & x^{ref} & 0 \end{bmatrix}^\top$ for a given final time $t_f$, define a one step cost 
\begin{equation}
 L_i^{ddp} = (X_i - X^{ref})^\top G_{ddp} (X_i - X^{ref}) + g_{ddp}  u_i^2, \nonumber
\end{equation}
where $G_{ddp}$ and $g_{ddp}$ are weights penalizing state deviation and control effort respectively. Then DDP can be used to generate a reference trajectory. This trajectory will be generated over the full horizon $t_f$ (or $N_{traj} = t_f / \Delta t$ steps), using the simplified model parameters $\lambda(q)$ at the initial state $q(0)$. 
Once we have the trajectory, $X^{traj}(i) = \begin{bmatrix} \theta^{traj}(i) & \dot\theta^{traj}(i) & x^{traj}(i) & \dot{x}^{traj}(i) \end{bmatrix}^\top$, it will be used as a time-varying reference for a receding horizong controller (or model predictive controller MPC) with a smaller horizon $t_H$ (or steps $N = t_H/\Delta t$) to generate closed loop control. Here the parameters of the simplified model $\lambda(q)$ will be updated at every time-step using the current state of the full robot. Define the one-step cost $L_i^{mpc} = (X_i - X_i^{traj})^\top G_{mpc} (X_i - X_i^{traj}) + g_{mpc} u_i^2$, the MPC scheme will generate a control sequence for the horizon $t_H$ by minimizing the cost over the horizon $t_H$. The first step of this control trajectory ($\ddot\theta^{ref}$) will be used as reference target for the low-level controller. The low-level controller will determine torques of all joints by minimizing a single step cost using quadratic programming as described earlier. This enables full-body 
participation in locomotion along with performing other manipulation tasks, such as end-effector orientation or gaze control.
\section{Results} \label{sec:res}
\begin{figure*}
 \centering
 \begin{subfigure}[h]{\textwidth}
  \centering
  \includegraphics[width=\textwidth]{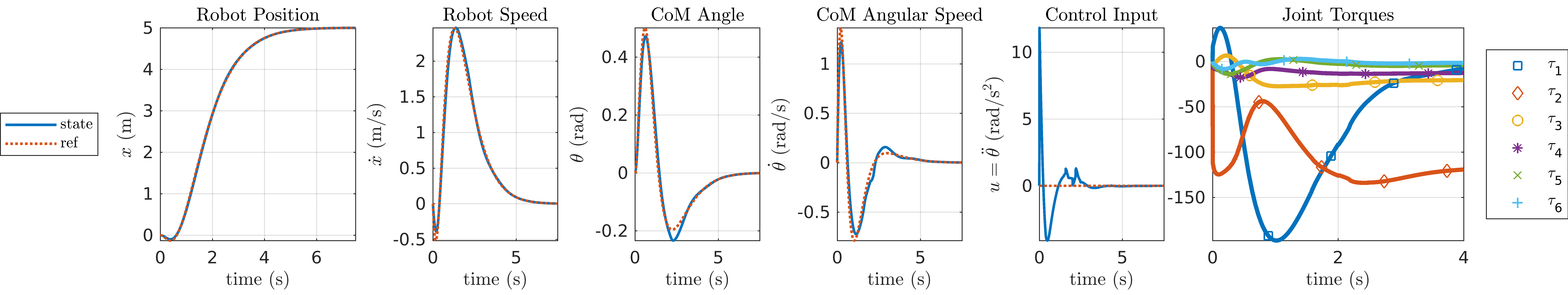}
  \caption{Simplified model reference and state trajectories (first 5 plots). \Munzir{The last plot shows the resulting joint torques for $t\in(0,4)$ seconds.}}
  \label{simple-result}
  \vspace*{0.5cm}
 \end{subfigure}
 \begin{subfigure}[h]{\textwidth}
  \centering
  \includegraphics[width=\textwidth]{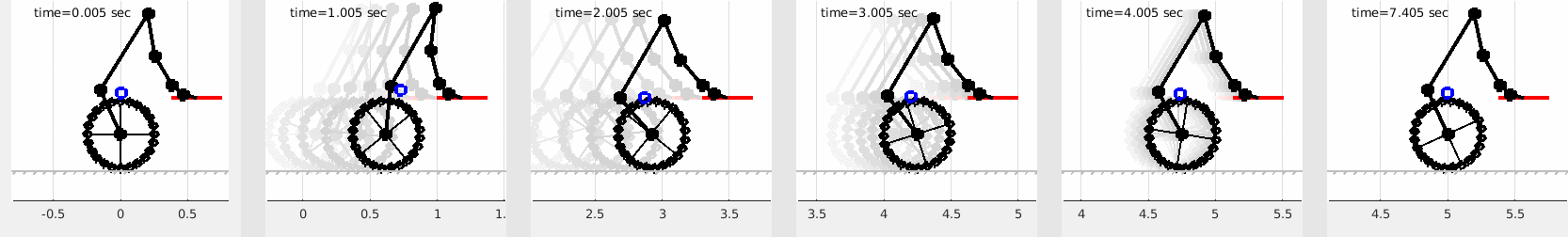}
  \caption{Snapshots of the full body at 6 different instances during execution. For each snapshot, body pose at five recent instances is also shown using shaded lines, to visualize speeds. Blue circle represents body CoM, and the red line represents an object attached to the last link at a fixed orientation. We see that the whole body is participating in manipulating the CoM to fulfill position objectives, while maintaining the orientation and position of the object attached to the end-effector.}
  \label{movie}
 \end{subfigure}
 \caption{}
\end{figure*}
We applied the presented approach on a simulation of 19-DOF robot (shown in Fig \ref{fig:frames}) with the objective of moving to a desired goal location on the ground while carrying a cup on a tray. The approach successfully manages to reach the goal location without letting the cup fall. We also make a comparison with the case of a traditional control \cite{stilman2014robots}, \ie without a unified control, to demonstrate its failure to prevent the cup from falling. The 3D simulation is uploaded as a video submission supplementing this paper. For the purpose of illustration, we provide here the details of applying the presented approach on a scaled down 7-DOF version of the same.

For trajectory generation using DDP, typically, a small penalty is assigned for each step and a large penalty is assigned for the terminal step. We have used a similar scheme. In our experiments, it turns out that for robots with large masses, a higher weight needs to be assigned to the pendulum angle $\theta(t)$. This ensures that the trajectory generated remains well within the stable region thus ensuring that the closed loop control (MPC) remains stable during execution. Note that MPC will not follow the trajectory generated by DDP exactly owing to the disturbances caused by full robot dynamics that were ignored in DDP. This means that $\theta$ may overshoot beyond the reference trajectory generated by DDP during execution. If $\theta$ was barely held within stable region, this may lead the full model to go unstable. For MPC, a horizon of 1 sec is used for each optimization step. A higher step cost is assigned to positions ($x(t)$ and $\theta(t)$) compared to speeds ($\dot{x}(t)$ and $\dot\theta(t)$). 
We have also used terminal weights for the MPC scheme, as they provide better tracking and stability performance. Terminal weights assign a high cost to the deviation of state at the end of the horizon at each control iteration. Final time $t_f = 20$ sec is used in the DDP to generate the results shown here. However the task was completed in $~7.5$ sec.

Finally, for the low-level controller, we have the following $P$ and $b$ matrices
\begin{equation}
\begin{aligned}[c]
P &= \begin{bmatrix} w_{\theta} J_{\theta} \\ w_{r_{ee}}J_{r_{ee}} \\ w_{\phi_{ee}}J_{\phi_{ee}} \\ w_{reg}\mathcal{I}_6 \end{bmatrix}  
\end{aligned}
\begin{aligned}[c]
b &= \begin{bmatrix} -w_{\theta}(\dot{J}_{\theta}\dot{q}-\ddot\theta^{ref}) \\ -w_{r_{ee}}(\dot{J}_{r_{ee}}\dot{q}-\ddot{r}_{ee}^{ref}) \\ -w_{\phi_{ee}}(\dot{J}_{\phi_{ee}}\dot{q}-\ddot\phi_{ee}^{ref}) \\ w_{reg}\ddot{q}_{reg} \end{bmatrix}
\end{aligned}
\end{equation}
For each task $x_t \in \{\theta, r_{ee}, \phi_{ee}\}$ the reference acceleration $\ddot{x}_t^{ref}$ appears in the defintion of $b$ and is defined as 
\[
 \ddot{x}_t^{ref} = \ddot{x}^d-K_p (x_t-x_t^d)-K_v(\dot{x}_t-\dot{x}_t^d)
\]
Desired position and orientation of the end-effector ($r_{ee}^d$ and $\phi_{ee}^d$) were set to their values at initial time. And the desired speeds and accelerations were set to zero. For pendulum angle $\theta$, the desired position and speed ($\theta^d$ and $\dot\theta^d$) come from the trajectory generated by DDP, while the desired acceleration $\ddot\theta^d$ is the control input determined by the MPC iteration. These reference values from DDP/MPC are used for the remaining sampling period ($0.01$ sec). During this period variable adaptive time-step simulation is used with low-level control QPs to simulate the behavior of the full robot.

Figure \ref{simple-result} shows the reference trajectory determined by DDP on the simplified model, and the state of the simplified model that results by applying the MPC control values on the full model. We see that the state follows the reference trajectory very closely with disturbances occuring during fast transitions owing to the disturbances caused by unmodeled full body dynamics. Also the last plot to the right is the plot of all joint torques. It is clear that the entire body is participating in performing the three tasks.

\section{Conclusion} \label{sec:con}
\Munzir{We presented a hierarchical approach to achieve whole body control of WIP humanoids
where a fast QP-based low-level controller generates body torques based on operational 
space objectives for manipulation and a closed-loop CoM motion policy generated
by a high-level controller. The latter utilizes MPC on a simplified model over a longer horizon. 
Time-varying reference for MPC is generated by DDP over a much larger horizon.
Equations of motion for a typical WIP humanoid helped with inverse dynamics QPs, and
for showing that body weight torque dominates wheel dynamics motivating the approximation
used for model predictions in the high-level controller. We demonstrated the approach on
a 19-DOF robot and presented a detailed account of a scaled down 7-DOF version of the same.} 

\small
\bibliographystyle{IEEEtran}
\bibliography{IEEEabrv,references}

\end{document}